# RoadSens-4M: A Multimodal Smartphone & Camera Dataset for Holistic Road-way Analysis

Amith Khandakar[1*], David Michelson[2], Shaikh Golam Rabbani[3], Fariya Bintay Shafi[3], Md. Faysal Ahamed[3], Khondokar Radwanur Rahman[3], Md Abidur Rahman[3], Md. Fahmidun Nabi[3], Mohamed Arselene Ayari[4], Khaled Khan[5], Ponnuthurai Nagaratnam Suganthan[5]

[1]*Department of Electrical Engineering, Qatar University, Doha, Qatar*
[2]*Department of Electrical and Computer Engineering, The University of British Columbia, British Columbia, Canada*
[3]*Department of Electrical & Computer Engineering, Rajshahi University of Engineering & Technology, Rajshahi 6204, Bangladesh*
[4]*Department of Civil and Environmental Engineering, Qatar University, Doha, Qatar*
[5]*Department of Computer Science and Engineering, Qatar University, Doha, Qatar*

**Corresponding Author*

**Amith Khandakar**

*Department of Electrical Engineering,*
*Qatar University,*
*Doha, Qatar*
Email Address: amitk@qu.edu.qa

# Abstract

It's important to monitor road issues such as bumps and potholes to enhance safety and improve road conditions. Smartphones are equipped with various built-in sensors that offer a cost-effective and straightforward way to assess road quality. However, progress in this area has been slow due to the lack of high-quality, standardized datasets. This paper discusses a new dataset created by a mobile app that collects sensor data from devices like GPS, accelerometers, gyroscopes, magnetometers, gravity sensors, and orientation sensors. This dataset is one of the few that integrates Geographic Information System (GIS) data with weather information and video footage of road conditions, providing a comprehensive understanding of road issues with geographic context. The dataset allows for a clearer analysis of road conditions by compiling essential data, including vehicle speed, acceleration, rotation rates, and magnetic field intensity, along with the visual and spatial context provided by GIS, weather, and video data. Its goal is to provide funding for initiatives that enhance traffic management, infrastructure development, road safety, and urban planning. Additionally, the dataset will be publicly accessible to promote further research and innovation in smart transportation systems.

# 1. Background & Summary

By 2040, the number of cars on the road is expected to double. This means that new ways to keep people safe on the road are needed. According to the World Health Organization (WHO), potholes and other road hazards can seriously damage automobiles and result in 20–50 million injuries and 1.3 million fatalities annually worldwide [1,2]. A 2014 AAA report says that potholes caused $6.4 billion in damage in the U.S. [3]. Researchers have used smartphone sensors like GPS and accelerometers, along with machine learning and wavelet decomposition, to find problems with roads in real time. But it's hard to move forward without standardised datasets and evaluation protocols. Some researchers have used simulation tools like CarSim (https://www.carsim.com/) to make fake data, while others have gotten real data from places like Chihuahua, Mexico. Unfortunately, many datasets are no longer accessible or lack the essential video and GIS context required for precise validation. This shows that we need a global benchmark dataset with both numbers and pictures to help us do better evaluations.

*Table 1. Survey of Prominent Datasets in the Field*

| Reference | Data Source | Environment Type | Public Access | Recorded Parameters |
|---|---|---|---|---|
| 4 | CarSim + Custom Test Data | Simulation-based | Public | Travel distance, velocity metrics, acceleration data, location tracking |
| 3 | Original collection | Field deployment | Private | Acceleration sensors |
| 1 | Original collection | Field deployment | Private | Acceleration + rotation sensors + positioning system |
| 5 | Pothole Lab platform | Simulation-based | Restricted | Acceleration measurements |
| 6 | Multi-source (Custom + Pothole Lab + CarSim) | Hybrid approach | Private | Position data, acceleration metrics, visual recordings |
| 7 | Pothole Lab + CarSim | Simulation-based | Private | Distance metrics, velocity, acceleration, positioning |
| 8 | Original collection | Field deployment | Restricted | Acceleration sensors |
| 9 | Multi-source (Custom + Pothole Lab + CarSim) | Hybrid approach | Private | Location tracking, acceleration, speed measurements |
| 10 | Original collection | Field deployment | Private | Acceleration data |
| 11 | Original collection | Field deployment | Public | Comprehensive sensor suite: acceleration (calibrated & raw), rotation (calibrated & raw), magnetic field (calibrated & raw), orientation, gravitational force, positioning system, combined acceleration |
| Current Work | Original collection | Field deployment | Public | Extended sensor suite: all measurements from Ref [11] plus visual road footage, meteorological conditions, geographic information system data |

AI and IoT improvements make it easier for cars to analyse sensor data when driving on their own, but finding road anomalies is still important for safe navigation. Recent multimodal datasets have improved techniques for finding anomalies. They have moved from image-based methods to multi-sensor fusion systems. Smartphone sensing looks like it could be useful on a large scale [12]. This survey identifies more than 25 important datasets, suggesting room for better integration of GIS data, smartphone sensors, and video , [13–16]. However, there are still gaps in effective multi-modal integration, especially when it comes to synchronising these different types of data.

Image-based road damage detection datasets like RDD2022 [16] and SVRDD [17] (see Table 2) have greatly improved automated infrastructure evaluation. RDD2022, which originated from RDD2020, expanded to include 47,420 images from six countries and over 55,000 identified instances of various damage types, such as potholes and longitudinal, transverse, and alligator cracks [16,18]. SVRDD [19] also has over 8,000 street-view images from Baidu Maps, with 20,804 marked damages. This makes it a useful tool for assessing roads on a city-wide scale. Multi-modal datasets like RSRD [20] and M2S-RoAD [20] combine stereo cameras, LiDAR, IMU, and RTK-GPS to make 16,000 high-resolution image pairs and labelled point clouds that can be used for self-driving applications. In addition, smartphone-based sensing datasets like the Comprehensive Smartphone Sensor Dataset [11] and the Irish Road Surface Condition Dataset [21,22] show that consumer-grade devices can be used for large-scale road monitoring and get good results even with low-power sensors. As shown in *Tables 1 and 2*, these datasets suggest that multi-source road anomaly detection is improving, but they still don't fully integrate across sensing, spatial, and environmental dimensions.

Even with these improvements, most of the datasets that are already out there don't have true multi-modal diversity and completeness. Even though RSRD [20] and M2S-RoAD [23] combine visual and depth modalities, they usually leave out synchronised video data, detailed Geographic Information System (GIS) layers, and weather information that is important for understanding real-world road conditions. Likewise, smartphone-based datasets [11,23] concentrate mainly on inertial sensing but are deficient in environmental context or spatial referencing, thereby constraining their utility for extensive infrastructure analytics.

*Table 2. An overview of the most advanced multimodal datasets*

| Dataset Name | Year | Publisher | Data Type | Modalities | Anomaly Types | Size | Geographic Coverage | Validation Method |
|---|---|---|---|---|---|---|---|---|
| RDD2022 [17] | 2022 | IEEE Big Data | Real-world | Smartphone cameras, drone imagery | Longitudinal/transverse/alligator cracks, potholes | 47,420 images, 55,000+ instances | Japan, India, Czech Republic, Norway, USA, China | CRDDC'2022 challenge, F1-score benchmarking |
| SVRDD [19] | 2024 | Scientific Reports | Real-world | RGB street view (1024×1024) | 6 crack types, potholes, patches, manholes | 8,000 images, 20,804 instances | Beijing, China (5 districts) | 10 algorithms tested, F1=0.709 |
| RSRD [20] | 2024 | Scientific Data | Real-world | Stereo cameras, LiDAR, IMU, RTK-GPS | Road geometry, elevation, macro-texture | 16,000 stereo pairs, point clouds | Multi-location | Depth estimation, stereo matching |
| Smartphone Sensors Dataset [11] | 2024 | Scientific Data | Real-world | Gravity, orientation, GPS, magnetometer, gyroscope, and accelerometer | Bumps, potholes, surface irregularities | 30+ km driving data, 89.82 Hz avg | Rajshahi, Bangladesh | Allan variance, T-tests, PCA, cross-device |
| TU-DAT [24] | 2025 | Sensors (MDPI) | Hybrid | CCTV, BeamNG simulation | Traffic accidents, aggressive driving | ~280 videos, 17,255 accident frames | Multi-location | Cross-dataset validation (CADP, DAD, AI-City) |
| URA-VLMs [25] | 2025 | Applied Sciences (MDPI) | Real-world | RGB images | Uneven surfaces, garbage, congestion, floods, fires | 3,034 annotated images | Urban scenarios | Multi-step prompting, RAG |
| KRID [26] | 2025 | Data (MDPI) | Real-world | RGB cameras | 34 infrastructure element types | Large-scale | Korea (highways, national, local roads) | Bounding boxes, polygon masks |

| Irish Road Dataset [27] | 2021 | Sensors (MDPI) | Real-world | Smartphone accelerometer (z-axis), GPS | Surface defects, potholes, rutting | 16 datasets, 30km route | Irish regional roads | K-means clustering, 84% accuracy |
|---|---|---|---|---|---|---|---|---|
| ISTD-PDS7 [28] | 2023 | Remote Sensing (MDPI) | Real-world | Natural CCD images | 7 distress types, 9 scenarios | 18,527 fine-grained labels | Multi-location | 7 segmentation models |
| UAV-PDD2023 [29] | 2023 | Data in Brief | Real-world | UAV cameras (30m altitude) | 6 types: cracks, potholes, repairs | 2,440 images, 11,158 instances | Multi-weather conditions | YOLO, Faster R-CNN |
| UAPD [30] | 2022 | Automation in Construction | Real-world | UAV high-res cameras | 6 categories: cracks, potholes, repairs | 3,151 images github | Various locations | Deep learning benchmarks |
| SmartRoadSense [31] | 2018-ongoing | Zenodo | Real-world crowdsensing | Smartphone accelerometer, GPS | Road surface roughness, quality | 75.9 MB aggregated | Global crowdsourcing | Crowdsensing validation |
| MobiLiteNet [32] | 2025 | Nature Communications | Real-world + synthetic | Smartphone/MR RGB (512×512) | 4 crack types, road markings | 2,873 original, 18,000 augmented | Germany, China, UK | 92.5% field accuracy, TensorFlow Lite |
| Autonomous Vehicle Dataset [33] | 2025 | IEEE DataPort | Simulated | Speed sensors, rotation sensors, GPS | False Data Injection Attacks | Comprehensive sensor suite | MOBATSim environments | Cybersecurity validation |

As shown in *Table 2*, even the most advanced multimodal datasets do not offer a unified fusion of all essential modalities. Our proposed dataset addresses these gaps by integrating mobile sensor data, synchronized video, GIS-based spatial mapping, and real-time weather attributes, creating a unified, context-rich resource for multimodal road anomaly analysis. This combination ensures greater diversity, completeness, and practical relevance, supporting the development of robust, real-world road monitoring and intelligent transportation systems.

## 2. Methods

In this part, we explain our plan for gathering and analysing data in detail. **Figure 1** shows our complete method, and the next subsections give more information about how we set up the experiment, chose the parameters, planned the route, collected data, and stored and protected it.

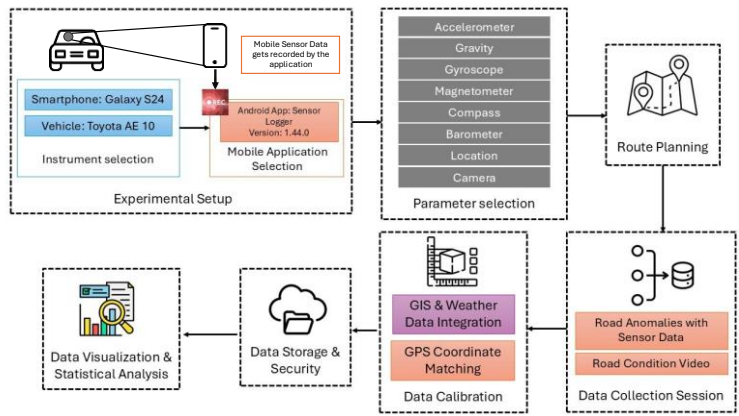

*Figure 1. An outline of the data collection process*

## 2.1. The Experimental Configuration

For this investigation, we collected data using a Samsung Galaxy S24 smartphone running Android 15.0 and the Sensor Logger program (version 1.46.0). Due to Android API limitations, we could only collect sensor data through the app, ensuring that sensor performance and data quality are consistent. We evaluated the noise characteristics of the accelerometer data under normal road conditions using Allan variance methodology, concentrating on the x, y, and z axes' random walk noise, flicker noise, and white noise. Acceptable thresholds were defined as 0.05–0.10 units for white and flicker noise, and 0.01–0.05 units for random walk noise. The results indicated all noise levels were within acceptable ranges: x-axis measurements showed white noise at 0.07 units, flicker noise at 0.07 units, and random walk at 0.02 units; y-axis measurements had white noise at 0.08 units, flicker noise at 0.08 units, and random walk at 0.04 units; and the z-axis displayed white and flicker noise at 0.05 units along with random walk at 0.02 units. This validation confirms that the accelerometer dataset is reliable and suitable for inertial measurement applications, as summarized in ***Table 3***. Furthermore, to ensure device-level consistency, cross-device evaluations were conducted between the Samsung Galaxy S24 and the Nothing Phone, showing similar accelerometer data patterns and comparable noise characteristics, as illustrated in ***Figure 2***.

*Table 3. Evaluation of the Data Collection Device's Noise Level and Allan Variance Results*

| Type of Noise | X | Y | Z | Acceptable Standards | Acceptance Results |
|---|---|---|---|---|---|
| **White Noise** | 0.07 | 0.08 | 0.05 | 0.05–0.10 | Pass |
| **Flicker Noise** | 0.07 | 0.08 | 0.05 | 0.05–0.10 | Pass |
| **Random Walk** | 0.02 | 0.04 | 0.02 | 0.01–0.05 | Pass |

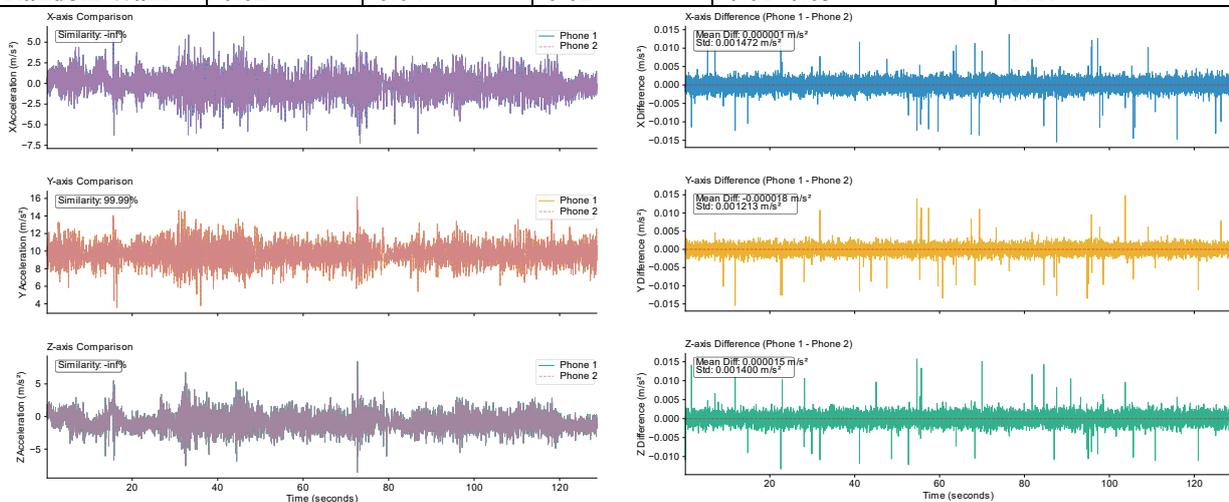

(A) *Accelerometer Cross Device Similarity Comparison*  (B) *Accelerometer Cross Device STD and Mean Difference*

*Figure 2. Comparison of accelerometer data patterns between the S24 and Nothing Phone, showing comparable trends and noise levels.*

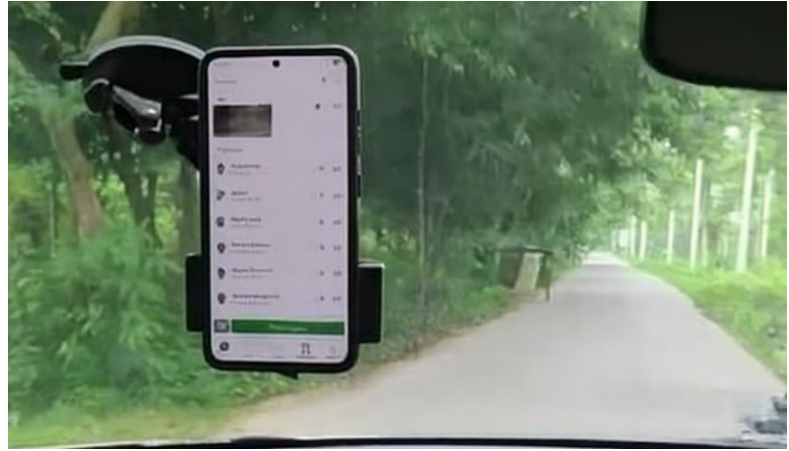

*Figure 3. Visual Guide to the Smartphone Installation Process*

We used a Toyota AE 110 to gather data, and we securely attached a smartphone to the windscreen with a high-quality suction mount. This setup ensured that the rear camera could see clearly and remain stable while collecting sensor data and recording video of the road conditions. The camera was in the middle of the car, which kept it steady and aligned with the vehicle's primary axes (x, y, and z). The suction mount's firm grip kept vibration to a minimum, even on rough surfaces. To make them more accurate, we put the sensors inside, where they wouldn't be as exposed to sunlight and temperature changes. Before conducting the experiments, we tested the setup on different types of roads to ensure it was stable. **Figure 3** shows how the installation on the windscreen was done safely, which was important for getting accurate sensor readings and reliable video of the road.

## 2.2. Parameter selection

Modern smartphones, like the Samsung Galaxy S24 used in this study, feature advanced sensors for capturing a variety of physical and environmental data. These sensors are divided into hardware-based, measuring physical phenomena directly, and software-based, which derive quantities from hardware data. The Sensor Logger app (version 1.46.0) was utilized to synchronize video and sensor streams, ensuring consistent timestamps. Key settings, as summarized in *Table 4* included a 'High' mode rear camera recording at 4K resolution and 30 FPS, with flash turned off to reduce lighting artifacts. GPS accuracy was kept under 0.001 degrees, and all motion sensors—accelerometer, gyroscope, magnetometer, and more—operated at over 100 Hz for detailed data collection.

*Table 4. Sensor Configuration Parameters*

|  | Parameter Category | Setting | Value/Description | Rationale |
|---|---|---|---|---|
| **Video Configuration** | Camera Selection | Rear Camera | Main rear camera with flash off. | Optimal road view without lighting interference |
|  | Image Quality | High | 4K Resolution with, 30FPS | Enhanced detail for road surface analysis |
|  | Recording Mode | Video | Continuous video recording | Synchronized with sensor data |
|  | Sampling Rate | Max | 56704 Kbps Bitrate | Optimal temporal resolution |
|  | Location Accuracy | Precise | <0.001 degree accuracy | High-precision georeferencing |
|  | Sensor Sampling | Max Frequency (>100Hz) | Device maximum for all sensors | Comprehensive data capture |
|  | Accelerometer | Max (>100Hz) | Hardware-dependent maximum | Detailed motion analysis |
|  | Gravity | Max (>100Hz) | Hardware-dependent maximum | Vertical force component isolation |

| Sensor Configuration | Barometer | Max (>1s) | Hardware-dependent maximum | Altitude and pressure variation tracking |
|---|---|---|---|---|
| | Compass | Max (>100Hz) | Hardware-dependent maximum | Directional heading determination |
| | Orientation | Max (>100Hz) | Hardware-dependent maximum | Device attitude and tilt measurement |
| | Gyroscope | Max (>100Hz) | Hardware-dependent maximum | Rotation rate measurement |
| | Magnetometer | Max (>100Hz) | Hardware-dependent maximum | Magnetic field variations |
| Data Logging Options | Calibrated Data | Enabled | Standard processed values | Immediate analysis capability |
| | Uncalibrated Data | Enabled | Raw sensor outputs | Custom processing flexibility |
| | Additional Streams | Disabled | Orientation/location extras off | Focused core measurements |
| | Data Streaming | Disabled | Single-device recording | Data consistency |

Data aggregation was optimized for fidelity and efficiency, with sensor readings resampled at 100 Hz and average aggregation applied to continuous signals. A null handling procedure was used to maintain authenticity. A simplified annotation system was in place for labeling anomalies like bumps and potholes, facilitating the development of a synchronized dataset that accurately records the visual and physical characteristics of imperfections in the road surface, as outlined in *Table 4*.

### 2.3. Route planning

The data collection took place in Rajshahi, a large city in Bangladesh, where specific areas with road issues, such as potholes and bumps, had already been identified. We made a detailed list of the longitude and latitude coordinates of each chosen location so that we could accurately target the anomaly sites. We made a structured route plan based on this catalogue so that we could systematically move through the designated areas. The selected route is illustrated in **Figure 4**, with red line marks indicating the precise path used to collect data. A bubble marks the starting point to indicate its location in Rajshahi. The last part of our route was 15.7 kilometres, and the whole route was 375.5 kilometres.

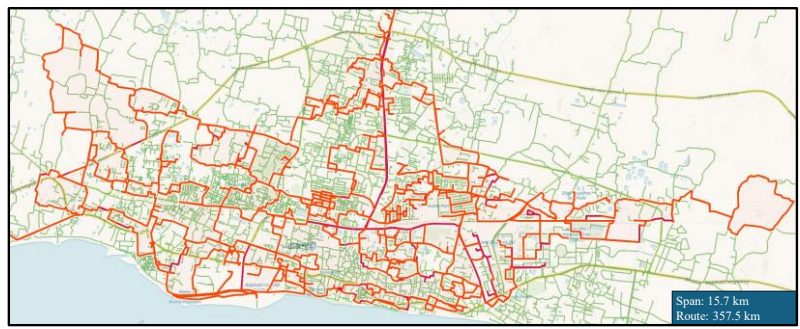

*Figure 4. The Planning of Routes (Red lines manually put to represent the data collection point)*

### 2.4. Mapping Smartphone Coordinates to Vehicle Reference Frame

Smartphone sensors usually show data values based on the device's screen orientation in a standard 3-axis coordinate system. When the smartphone is held horizontally, the z-axis is at a right angle to the screen, the x-axis runs from left to right, and the y-axis runs from bottom to top. During the data collection process, the coordinate system of the smartphone and the vehicle were meticulously aligned, as seen in **Figure 5**. The smartphone's x-axis aligns with the vehicle's longitudinal axis, its y-axis aligns with the lateral axis, and its z-axis aligns with the vertical axis. Therefore, the force of gravity in relation to the Earth's coordinate system is well represented by the z-axis of the vehicle's coordinate system. The sensors on the smartphone needed to be perfectly aligned to provide accurate information about how the car was moving and changing.

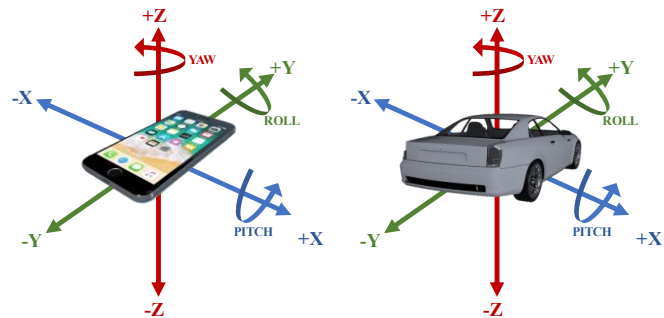

*Figure 5. Alignment of the device and vehicle coordinates*

All motion sensors transmit multidimensional arrays of data when a sensor event occurs. The gyroscope measures the speed at which an object rotates along all three axes, whereas the accelerometer measures the speed of motion along the x, y, and z axes. The gravity sensor provides a 3D vector indicating the direction and magnitude of gravity. The magnetometer detects changes in the magnetic field. It measures the raw magnetic field strength along all three axes, while the orientation sensor records the device's position with respect to magnetic north. This sensor data, along with additional details about each sensor event, can be obtained as float arrays. This type of information enables an exact analysis of the vehicle's motion and the condition of the route.

## 2.5. Data collection session

The process of collecting data took eight days and covered more than 375 kilometers of roads in cities and suburbs. As shown in **Figure 4**, where red lines indicate the data paths through various road conditions, careful route planning ensured that different road irregularities, such as bumps and potholes, were accounted for. We focused on stable weather and GPS accuracy, keeping track of these things for later use to compare sensor data with things like light and temperature. The motion values were verified by aligning the smartphone's coordinate system with the vehicle's axes, as illustrated in **Figure 5**. The x-axis was the same as the vehicle's long axis, the y-axis was the same as the vehicle's lateral axis, and the z-axis was the same as the vehicle's vertical axis. This made it possible to record acceleration and orientation accurately. The Sensor Logger app (version 1.46.0) used optimal settings for recording 4K video at 30 frames per second, achieving GPS accuracy within 0.001 degrees, and sampling sensors at rates over 100 Hz.

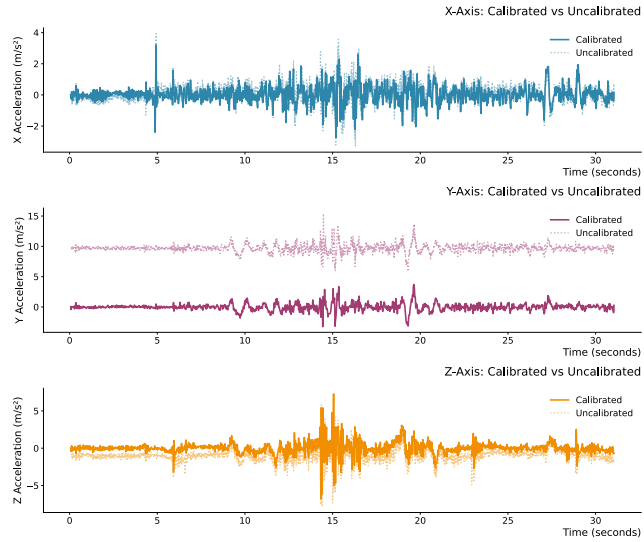

*Figure 6. Comparison of the Accelerometer's Calibrated and Uncalibrated Data*

Both calibrated and uncalibrated data were recorded simultaneously, with calibration significantly improving signal consistency, especially along the y-axis, as seen in ***Figure 6***. A semi-manual annotation system was implemented for precise labeling. Observers pressed the annotation button at the start and end of an anomaly, ensuring consistent labeling across the dataset. The Sensor Logger app synchronized video recordings and sensor data with identical timestamps, allowing for a unified dataset capturing both visual and inertial data of road anomalies. This strategy facilitated high-quality data collection suited for model training and validation.

## 2.6. Video-Sensor Synchronization

We utilized the Sensor Logger app (v1.46.0) to synchronize timestamps on video recordings and sensor data using the device's internal clock for millisecond-resolution accuracy. This method allows for direct comparison of sensor readings with visual events, resulting in an exported dataset that contains raw video files and an annotation CSV listing the start and end times of detected road issues, such as bumps and potholes. **Figure 7** shows a typical frame from an event when someone hit a pothole.

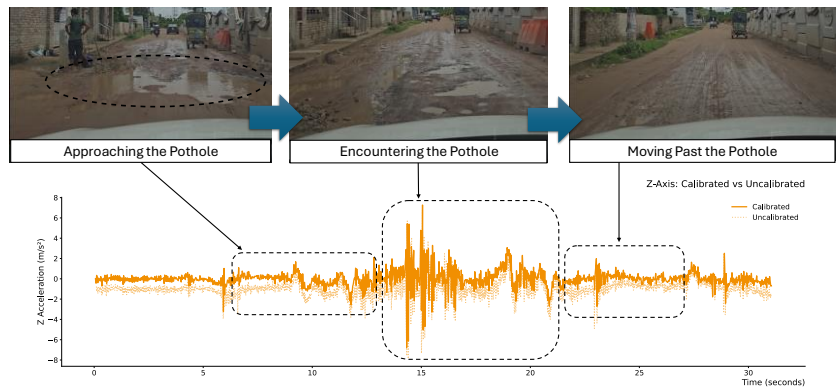

*Figure 7. Snapshot of the Road Condition Video Recorded on a Pothole Road Anomaly*

To ensure accurate synchronization, we conducted cross-modal calibration with controlled tests that captured known events in both video and sensor channels, confirming precise alignment between sensor spikes and visual cues. The Samsung Galaxy S24's hardware timestamping maintains synchronization among all sensors and the camera. The annotation file acts as a ground-truth reference, enabling systematic comparison of sensor anomalies against video evidence and supporting analysis and algorithm development. **Figure 7** illustrates the alignment of video data with accelerometer sensor data, demonstrating proper synchronization between annotations, video, and numerical data.

## 2.7. Post-Processing, GIS, and Weather Integration

In the post-processing phase, various contextual data sources were integrated to increase the precision and interpretability of sensor data. The RouteDoodle platform was used to map the data collection path, which was roughly 357.5 kilometers across Rajshahi, Bangladesh, providing detailed elevation profiles. Elevation data were aligned with sensor coordinates to ensure accuracy, and a 3D terrain model was created **(Figure 8A)**. Elevation distributions and profiles **(Figures 8B–D)** illustrated terrain variations, aiding analysis of how vertical gradients impact accelerometer and gyroscope readings. To enhance the dataset, GIS data for the Rajshahi area was obtained from NextGIS, allowing for in-depth spatial analysis compatible with QGIS. GIS elevation data were merged with RouteDoodle elevations and sensor readings, creating a unified dataset that links timestamped sensor data with spatial attributes, supporting the examination of road geometry and vehicle dynamics.

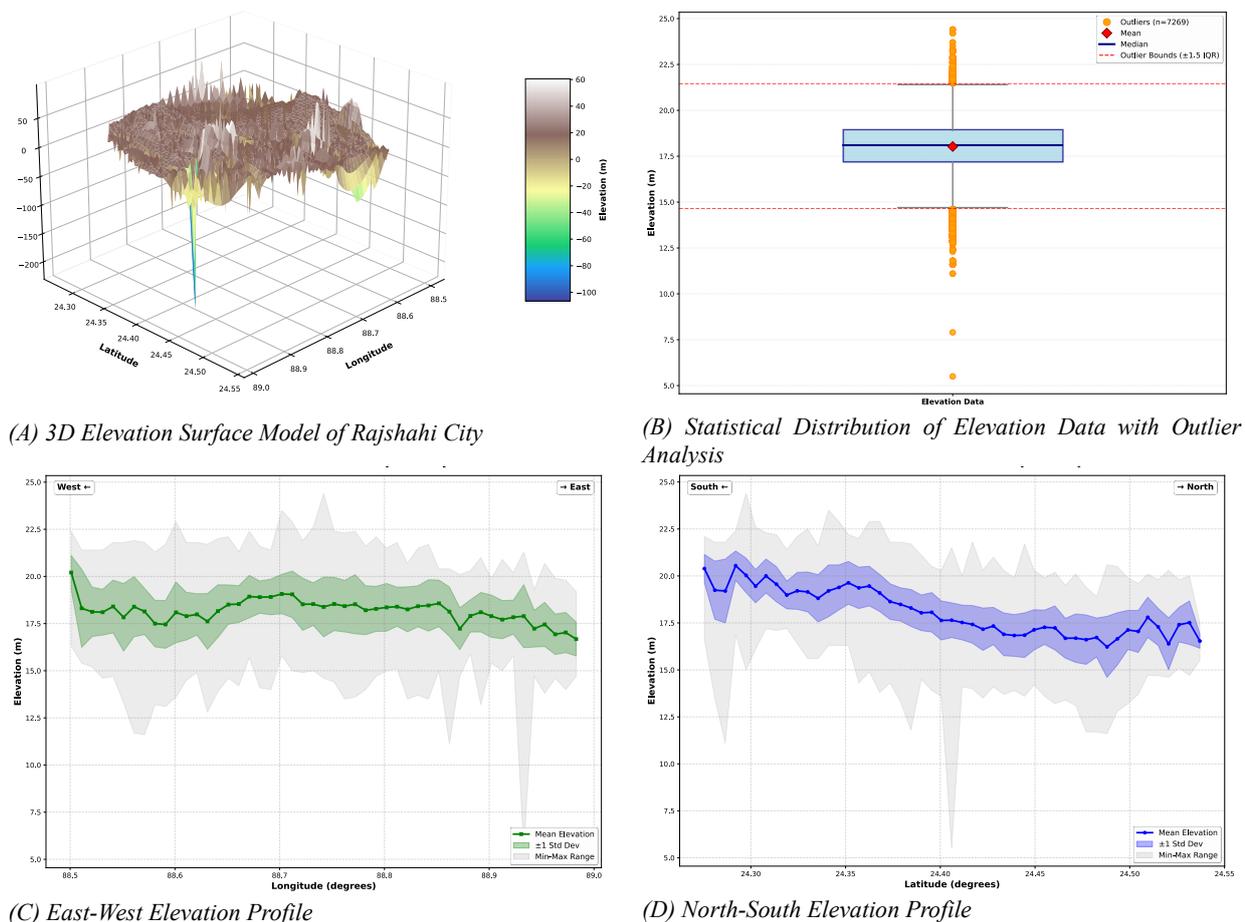

*(A) 3D Elevation Surface Model of Rajshahi City*

*(B) Statistical Distribution of Elevation Data with Outlier Analysis*

*(C) East-West Elevation Profile*

*(D) North-South Elevation Profile*

**Figure 8. GIS and Weather Information Graphs of the Comprehensive Dataset: (A) 3D Elevation Surface Model of Rajshahi City, (B) Statistical Distribution of Elevation Data with Outlier Analysis, (C) East-West Elevation Profile, (D) North-South Elevation Profile**

Weather data from the Sensor Logger app—logging temperature, humidity, cloud coverage and atmospheric pressure—was also incorporated to ensure environmental consistency. To reduce interference, recordings were made between 10:00 AM and 1:00 PM in clear weather. The final dataset includes synchronized sensor readings, video footage, GIS coordinates, elevation, and weather information, offering a comprehensive resource for studying road surface dynamics in real-world conditions.

## 2.8. Calibration of sensor data

To obtain more accurate measurements, known as calibrated data, raw data from smartphone sensors must be adjusted to account for any errors or biases. Uncalibrated data, on the other hand, is the sensor's original output without any changes. This could be useful for specialized processing.

The Sensor Logger Android app can get data from accelerometers, magnetometers, and gyroscopes, whether they have been calibrated or not. Internal calibration fixes problems like bias (when the sensor keeps giving you wrong information) and sensor drift (when the sensor changes slowly over time). For example, when an accelerometer is stationary, its readings should be 0 m/s² on the x- and y-axes and approximately 9.8 m/s² on the z-axis. There is bias in any departure. The software calibrates sensors by finding consistent deviations when they should read zero. For example, a gyroscope should read −0.43 rad/s when it is not moving. You can use calibration to fix this bias. The next thing to do is calibrate the accelerometer:

$$C_a = S_a - O_a \qquad\qquad 1$$

Where $C_a$ is the output of the calibrated accelerometer, $S_a$ is the raw accelerometer measurement, and $O_a$ is the offset calculated during calibration. To reduce bias-related mistakes, the accelerometer's x, y, and z values should ideally read zero when the device is stationary after calibration. ***Table 5*** shows the accelerometer's offsets and calibrated values. The app uses both calibrated and uncalibrated sensor readings for the gyroscope and magnetometer, which are visualized in ***Figure 10 (F, G, H, I)***.

*Table 5. Offsets and Calibrated Values for Accelerometer*

| Parameter | X-axis | Y-axis | Z-axis |
|---|---|---|---|
| Raw Offset | $\mathbf{0.410\ m/s^2}$ | $0.427\ m/s^2$ | $9.814\ m/s^2$ |
| Corrected Mean | $-0.176\ \mathbf{ms^2}$ | $-0.428\ \mathbf{ms^2}$ | $-0.059\ \mathbf{ms^2}$ |

**Figure 6** illustrates the distinctions between calibrated and uncalibrated accelerometer data in the bump category. For the X, Y, and Z axes, the corresponding offsets were 0.410, 0.427, and 9.814 m/s². -0.176 m/s², -0.428 m/s², and -0.059 m/s² were the calibrated mean values while the item was stationary. Calibration successfully removed biases from sensor readings, as shown in Table 5. This brought the readings closer to what was expected. The magnetometer is also calibrated using an ellipsoid fitting method, which starts with raw data collected at different angles to fix distortions caused by hard and soft iron. This is how the dataset is made:

$$S_m = (x, y, z) \qquad\qquad 2$$

To appropriately depict the magnetic field, the data is fitted to an ellipsoid, whose general equation is as follows:

$$\left(\frac{x - h_x}{a}\right)^2 + \left(\frac{y - h_y}{b}\right)^2 + \left(\frac{z - h_c}{c}\right)^2 = 1 \qquad\qquad 3$$

Here, $(h_x, h_y, h_z)$ symbolizes the ellipsoid's center, while a, b, and c are its radii along the corresponding axes. By fitting the ellipsoid, biases in the raw magnetometer data can be found and fixed, improving the output's reflection of the actual magnetic field. The magnetometer calibration is then performed using the following formula

$$C_m = S_m - O_m \qquad 4$$

$C_m$ is the output of the calibrated magnetometer, $S_m$ is the raw magnetometer measurement, and $O_m$ is the offset that was discovered during ellipsoid fitting. In more complicated situations, it may be necessary to make both offset corrections and scaling changes. This can be said as:

$$M = K \cdot (B - O) \qquad 5$$

M is the calibrated output, B is the uncalibrated magnetometer measurement, O is the determined offset, and K is the calibration gain or scaling factor. During the calibration process, external factors such as misalignment or errors in manual positioning can introduce additional bias. For example, if the device is accidentally tilted or rotated during calibration, the sensor readings may be slightly inaccurate, but they will remain consistent. To minimize these errors, the calibration process requires that the device be placed on a stable, level surface and perfectly aligned.

Over time, sensors can drift due to changes in the environment, wear and tear, or modifications to internal parts. This can cause them to be wrong even after they have been calibrated. Recalibrating the sensor on a regular basis helps to reduce bias caused by this drift, making sure that the readings are correct. There may still be random noise, but filtering methods can help with this. Calibrated data gives you instant accuracy, which is necessary for accurate measurements. Uncalibrated data, on the other hand, gives you a detailed view of raw outputs, which is helpful for custom calibration or deeper analysis. This study provides both calibrated and uncalibrated data for comprehensive analysis and accurate measurements.

### 2.9. Storage Architecture and Data Security

The data was saved in a CSV file with 12 columns, each containing data from a different sensor, such as an accelerometer, gyroscope, or GPS. A metadata file had timestamps and environmental factors, while an annotation file kept track of road events like bumps and potholes. We combined all the sensor data and notes into one CSV file, which was divided into 0.01-second chunks for analysis. The Sensor Logger app used the mode for non-numeric values and averaged numeric attributes to downsample. The "absolute before downsampling" option was off, and no upsampling was done to keep the original signal distribution. A video folder linked recordings of road problems to sensor readings to check their accuracy. Google Cloud Storage was used to store regular backups to keep data safe.

# 3. Data Structure

## 3.1. Structure

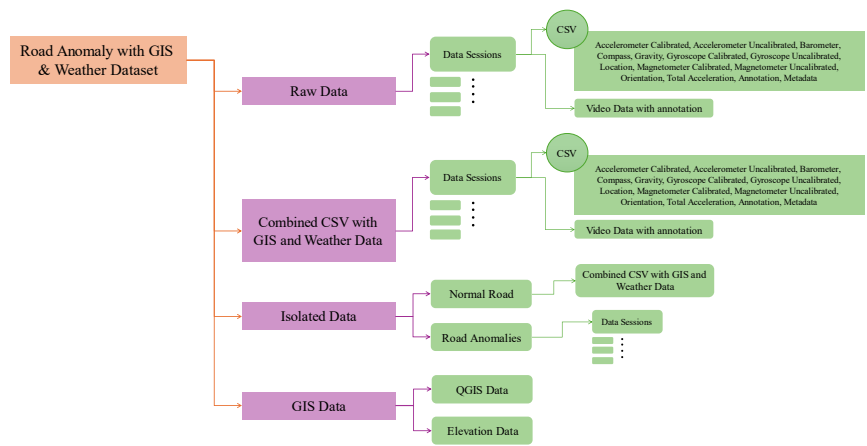

*Figure 9. Directory Structure and File Organization of the Dataset*

The Road Anomaly with GIS and Weather Dataset is set up in a way that makes it easy to access and combine with other types of data. Every session folder has all of the sensor CSV files, as well as annotation and metadata files with readings from the accelerometer, gyroscope, magnetometer, barometer, compass, and GPS modules that have been calibrated and not calibrated. There is a separate camera subfolder in each session that holds the annotation file and a text file with links to the video data on Google Drive. This structure makes it easy for researchers to get to full video recordings without having to manually extract clips. All of the annotated video files are in the Google Drive folder, along with the start and end times for each road anomaly that was found.

There are four main parts to the dataset: Raw Data, Combined CSV with GIS and Weather Data, Isolated Data, and GIS Data. The Combined CSV section combines sensor readings with information about the environment and location, such as temperature, humidity, and geographic coordinates. The Isolated Data folder keeps normal road and anomaly samples apart so they can be analysed more closely. The GIS Data folder has QGIS files and elevation data that can be used for spatial mapping and visualisation. This well-organised structure makes sure that research applications are consistent, modular, and easy to use. **Figure 9** shows how the dataset's files and directories are set up. The dataset is publicly available on Figshare at https://doi.org/10.6084/m9.figshare.30341143.v2 [34].

## 3.2. Description of fields

There are several CSV files in the dataset, each containing a different type of sensor data collected for the study. The following are the six files: TotalAcceleration.csv, Gyroscope.csv, GyroscopeUncalibrated.csv, Accelerometer.csv, AccelerometerUncalibrated.csv, and Gravity.csv. Some fields present in every file include Time (ns), representing the Unix timestamp of the data point, and Seconds elapsed (s), indicating the time in seconds since the data collection started. The dataset also includes acceleration fields for the x, y, and z axes (in microtesla for magnetometers or meters per second squared for accelerometers), labeled as x (m/s² or µT), y (m/s² or µT), and z (m/s² or µT). The Orientation.csv file keeps track of the device's 3D orientation by using rotation angles (Roll (rad),

Pitch (rad), and Yaw (rad)) and quaternion components (Qx, Qy, Qz, and Qw). The angles show how the device is oriented in relation to the Earth's axes.

The dataset also has a Combined.csv file that puts all the sensor data into one format. Each row is lined up with the same timestamp from all the sensors, and there is always a 0.01-second time difference between records. A GIS and Weather folder has been added to the dataset to enhance its quality. This folder contains a QGIS data file for those interested in the terrain or elevation of the city where the data was collected. It also has a Weather file that shows the weather conditions for the days when the data was collected. Lastly, the dataset has a Video folder that stores video data showing how the roads were during the time the data was collected. This video footage serves as a visual reference to accompany the sensor data, providing more information to help you identify road problems and driving habits. *Table 6* provides a comprehensive list of the dataset's contents, including the number of records, sensors, and data types.

*Table 6. Dataset Summary Statistics*

| Category | Volume | Files | Remarks |
| --- | --- | --- | --- |
| Combined Dataset (Total) | 1,034,526 rows | 103 files | GIS + weather integrated |
| Normal Road Baseline conditions | 44,047 rows | 5 files | Pure normal conditions |
| Mixed Road Survey | 990,479 rows | 98 files | Contains both normal |
| Video Collection | 10,300 seconds | 103 files | ~2.86 hours |
| GIS Elevation Data | 483,866 rows | - | Geospatial data |

# 4. Data Validation and Quality Assurance

The data was stored in a tabular CSV format, comprising 12 different CSV files for various sensors, such as GPS, accelerometers, gyroscopes, and more. A metadata file kept track of important information like timestamps and environmental factors, while an annotation file kept track of road events like bumps and potholes. To make it easier to analyse, all of the sensor data and notes were put into one CSV file, which was split into 0.01-second intervals. The Sensor Logger app used the mode for non-numeric attributes to keep the most common values and averaged numeric attributes for downsampling. The "absolute before downsampling" option was not on, and there was no upsampling to keep the original signal distribution.

A folder of videos showing road problems was also made, and these videos were linked to the sensor readings for cross-modal validation. We made regular backups on Google Cloud Storage to protect the data for future research and validation.

## 4.1. Road Anomaly Detection

To make sure the dataset was correct and trustworthy, sensor signals from road problems like bumps and potholes were carefully checked. This included visually examining the data and performing statistical analysis, which involved calculating the mean, variance, and standard deviation, as well as conducting a T-test to determine if the results were significant. We also did a cross-correlation analysis to look into the link between bump and pothole signals. The dataset's reliability and integrity for use in vehicle dynamics and road safety research and applications are confirmed by this extensive validation.

## 4.2. Visualization of data plots

During quality control, we often plotted sensor signals to ensure the readings were accurate. To find any problems, it was important to monitor how well the sensors were working. The accelerometer can detect when a car's speed changes quickly due to bumps and potholes in the road. A bump, for example, usually makes the acceleration increase quickly, but a pothole causes it to decrease quickly. These strange things could also alter the caris orientation, potentially causing spikes or changes in the gyroscope data. Sudden changes in altitude can also affect GPS readings.

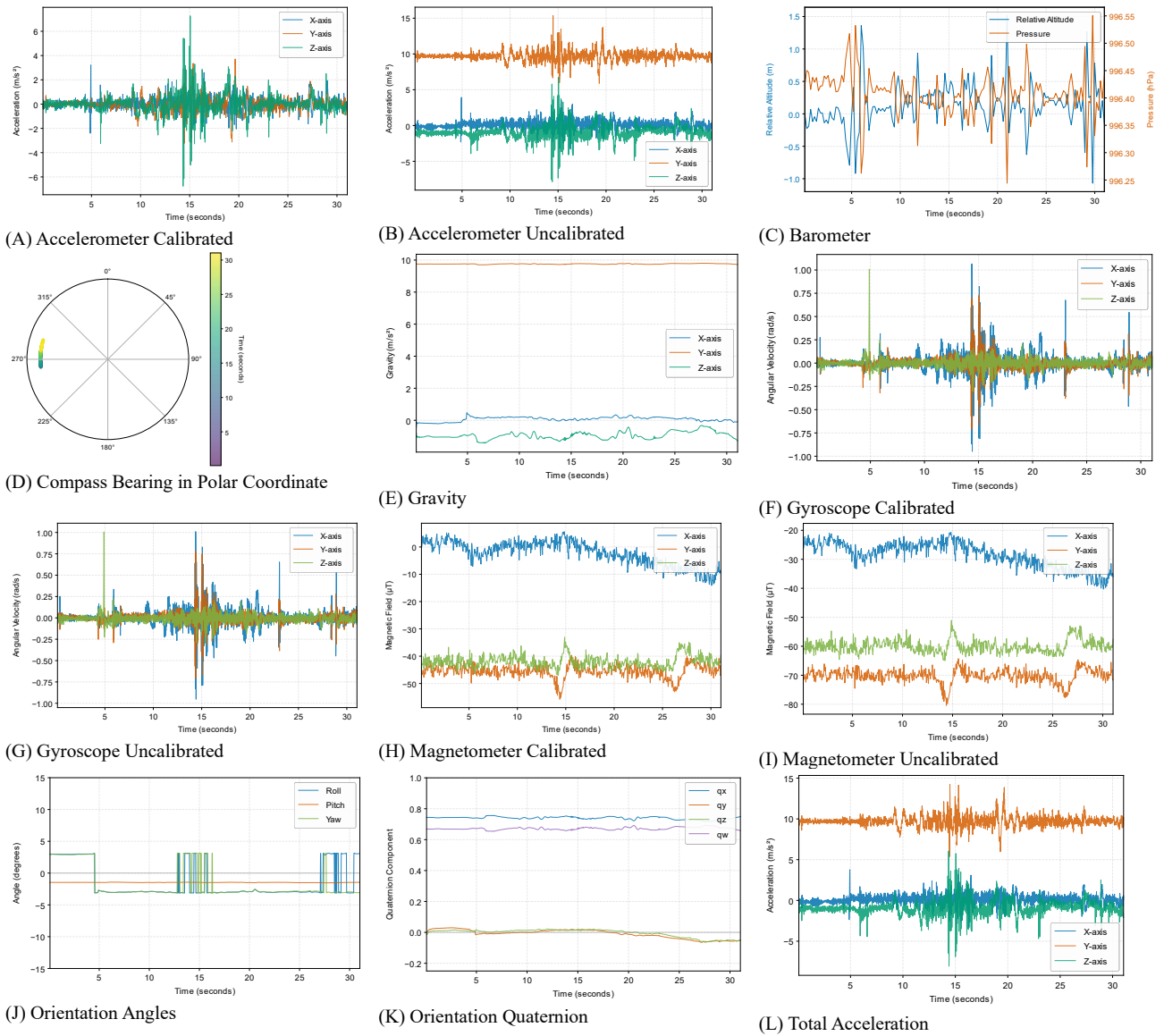

(A) Accelerometer Calibrated  (B) Accelerometer Uncalibrated  (C) Barometer
(D) Compass Bearing in Polar Coordinate  (E) Gravity  (F) Gyroscope Calibrated
(G) Gyroscope Uncalibrated  (H) Magnetometer Calibrated  (I) Magnetometer Uncalibrated
(J) Orientation Angles  (K) Orientation Quaternion  (L) Total Acceleration

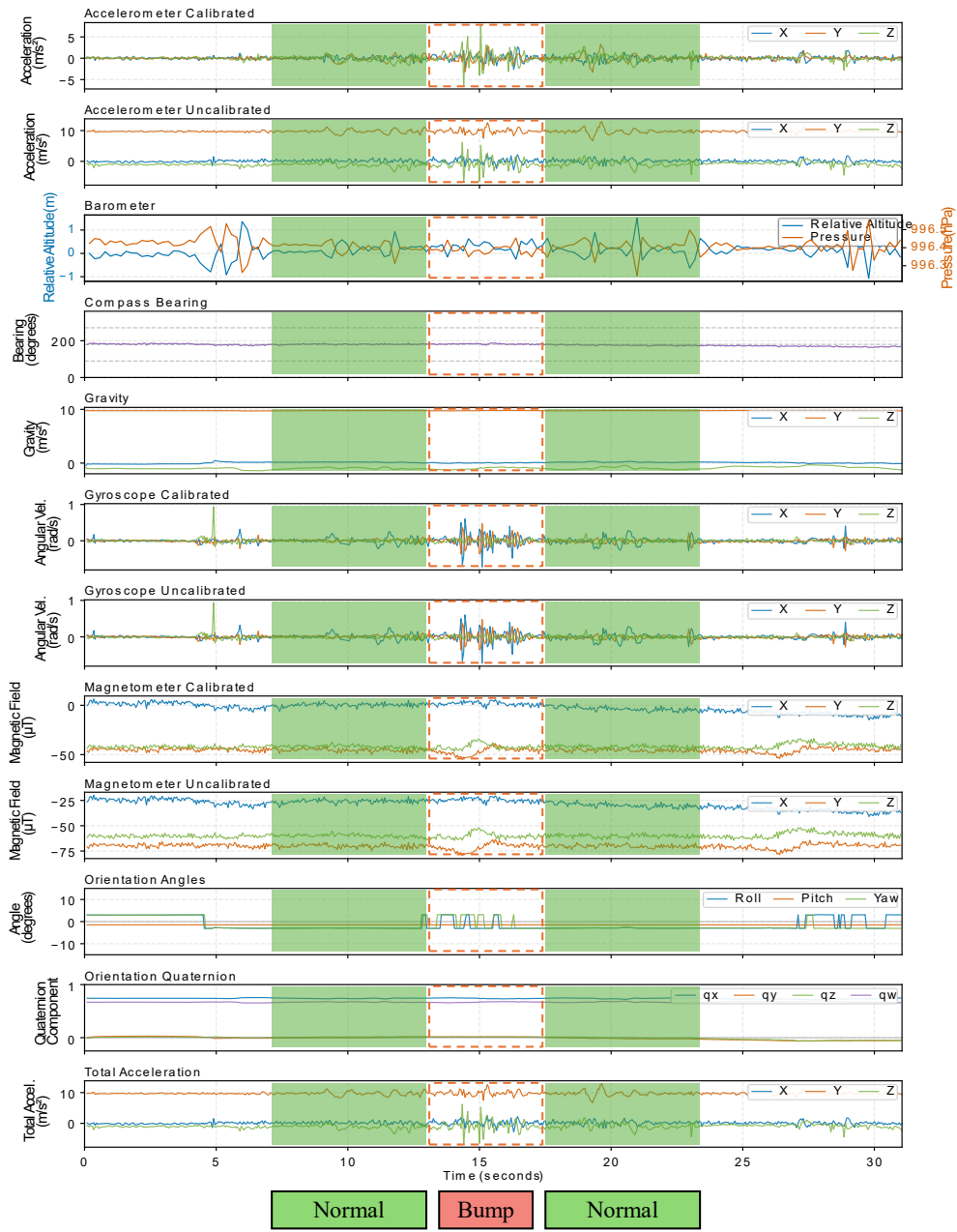

(M) Combined CSV with 0.01 Second Down sampling for Time Alignment

*Figure 10. Comprehensive Sensor Data Overview of a Driving Session (A) Accelerometer Calibrated, (B) Accelerometer Uncalibrated, (C) Barometer, (D) Compass, (E) Gravity, (F) Gyroscope Calibrated, (G) Gyroscope Uncalibrated, (H) Magnetometer Calibrated, (I) Magnetometer Uncalibrated, (J) Orientation, (K) Orientation Quaternion (L) Total Acceleration, (M) Combined CSV with 0.01s Downsampling*

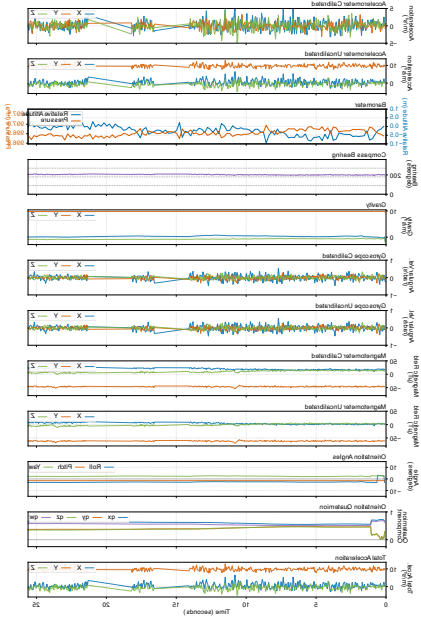
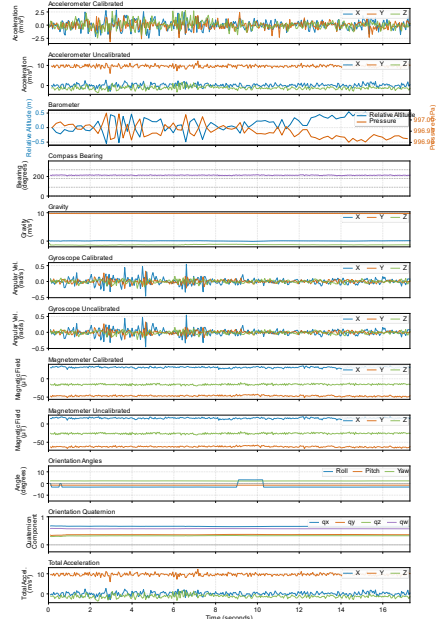
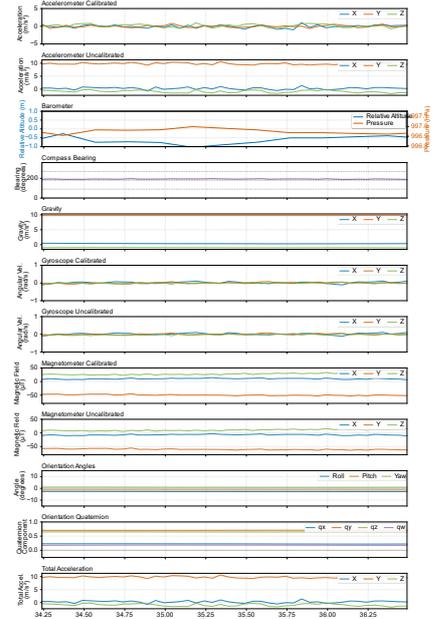

*Figure 11. Sensor Signal Response During a Bump Event*

*Figure 12. Sensor Signal Response During a Pothole Event*

*Figure 13. Sensor Signal Behavior During Normal Road Conditions*

**Figures 10, 11, 12**, and **13**, located in Folder 3 within the Non-Isolated folder, illustrate the sensor signals under various road conditions. **Figure 10** provides a comprehensive overview of the complete driving session data for all sensors, including both calibrated and uncalibrated accelerometer readings (*A* and *B*), barometer data (*C*), compass readings (*D*), gravity data (*E*), calibrated and uncalibrated gyroscope data (*F* and *G*), calibrated and uncalibrated magnetometer readings (*H* and *I*), orientation data (*J*), orientation quaternion (K), total acceleration (*L*), and combined CSV data with 0.01-second downsampling (*M*). **Figure 11** captures the sensor signal response during a bump event, highlighting the distinctive patterns across multiple sensor modalities. In contrast, **Figure 12** presents the sensor signal response during a pothole event, demonstrating how sensors react differently to potholes compared to bumps. Lastly, **Figure 13** shows the sensor signal behavior under normal road conditions, providing a baseline for comparison with anomalous events. Together, these plots give a detailed view of how the sensors respond to different road surfaces, enabling clear differentiation between normal driving conditions and road anomalies.

## 4.2. Statistical analysis

Z-axis accelerometer data was statistically analyzed under three different driving conditions: bumps (1,635 samples), potholes (242 samples), and normal roads (1,726 samples). Descriptive statistics indicated key differences: bump data had a mean of 0.045 m/s² with high variability ($\sigma$ = 1.096 m/s²) and a range of 14.214 m/s², pothole data had a negative mean of -0.227 m/s² and the highest variance ($\sigma^2$ = 1.397 (m/s²)²). In comparison, normal road data showed the least variability ($\sigma$ = 0.817 m/s²) with a near-zero mean (0.011 m/s²) (see ***Table 7***). One-sample t-tests reinforced these findings, showing significant negative bias for pothole data (t = -2.992, p = 0.003) but no significant deviation for bumps and normal conditions. Distribution analysis revealed that bump data had a slight positive skewness (0.418) and heavy tails (kurtosis = 7.700), pothole data had a slightly negative skewness (-0.306) with mesokurtic traits (kurtosis = -0.157),. Normal road data had positive skewness (0.366) with moderately heavy tails (kurtosis = 2.001). Independent t-tests indicated significant differences between bump and pothole (t = 3.573, p < 0.001) and between pothole and normal road (t = -3.987, p

< 0.001), but not between bumps and normal roads (t = 1.039, p = 0.299). One-way ANOVA confirmed at least one significant mean difference (F = 8.168, p < 0.001) (see *Table 7*).

Table 7. Statistical Analysis of Accelerometer Z-Axis Data of Pothole and Bump and Combined Raw CSV

| Metric | Isolated Bump CSV | Isolated Pothole CSV | Raw Sensor CSV (Normal) |
|---|---|---|---|
| Sample Size (n) | 1,635 | 242 | 1,726 |
| Mean (μ) | 0.045182 m/s² | -0.227365 m/s² | 0.010668 m/s² |
| Variance (σ²) | 1.201703 (m/s²)² | 1.397386 (m/s²)² | 0.666918 (m/s²)² |
| Standard Deviation (σ) | 1.096222 m/s² | 1.182111 m/s² | 0.816651 m/s² |
| Range | 14.213785 m/s² | 6.499470 m/s² | 8.233954 m/s² |
| T-Statistic (one-sample) | 1.666587 | -2.992077 | 0.542688 |

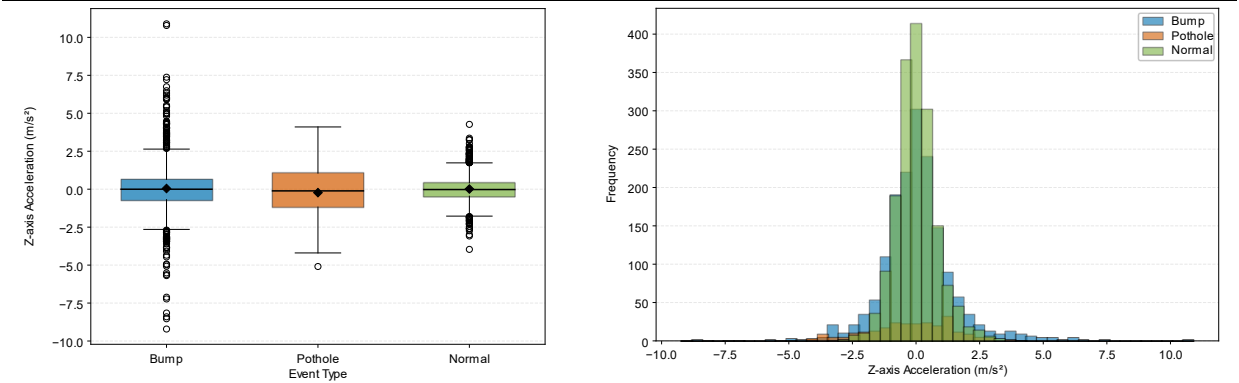

*Figure 14: Using z-axis accelerometer data, statistical evaluations of bump versus pothole Boxplot analysis (A) shows the variance and distribution of acceleration data for potholes and bumps, while histogram analysis (B) shows the frequency distribution of z-axis accelerations for each type of event.*

Cross-correlation analysis showed strong autocorrelation for bump data (0.681) and moderate for potholes (0.619); normal road data had the lowest (0.430). As illustrated in **Figure 14**, the boxplot (A) highlights the variance and distribution contrasts between bumps and potholes, while the histogram (B) visualizes the frequency distributions of z-axis accelerations for each event type—showing overlapping yet distinguishable profiles. Histograms demonstrated overlapping yet distinguishable distributions, while boxplots illustrated increasing variability from normal to bump conditions. Overall, the analysis confirms unique accelerometer signatures for each road condition, providing a solid basis for effective road condition classification and severity assessment.

### 4.3. Multivariate sensor analysis for road anomaly detection

A statistical analysis was performed on z-axis accelerometer data collected under various road conditions, including 1,635 samples of bumps, 242 samples of potholes, and 1,726 samples of normal road surfaces. The descriptive statistics, summarized in *Table 7*, revealed the following findings: bump data had a mean acceleration of 0.045 m/s², accompanied by high variability (σ = 1.096 m/s²) and a wide range (14.214 m/s²). Pothole data exhibited a mean of -0.227 m/s² and the highest variance (σ² = 1.397 (m/s²)²), while normal road data showcased the lowest variability (σ = 0.817 m/s²) and a near-zero mean (0.011 m/s²). One-sample t-tests indicated that the means for bumps and normal road data did not significantly deviate from zero. At the same time, potholes displayed a significant negative bias (t = -2.992, p = 0.003), confirming distinct acceleration patterns for different road conditions.

Further analysis of the distributions revealed differences in skewness and kurtosis among the road types: bump data had a positive skewness (0.418) and heavy tails (kurtosis = 7.700), pothole data showed slight negative skewness (-0.306) with mesokurtic characteristics, and normal road data exhibited slight positive skewness (0.366). Independent t-tests indicated significant differences between bump and pothole data (t = 3.573, p < 0.001) and between pothole and normal road data (t = -3.987, p < 0.001). However, there was no discernible difference between the data from bumps and regular roads. (t = 1.039, p = 0.299). One-way ANOVA confirmed at least one significant mean difference (F = 8.168, p < 0.001), supporting the descriptive trends displayed in *Table 7*.

*Table 7. Statistical Overview of the Main Elements by Condition of the Road*

| Component | Condition | n | Mean (μ) | Std Dev (σ) | Min | Max | Range |
|---|---|---|---|---|---|---|---|
| PC1 | Bump | 1,635 | 3.631 | 0.563 | 1.461 | 5.385 | 3.924 |
| | Pothole | 242 | -1.749 | 0.435 | -2.662 | -0.616 | 2.046 |
| | Normal | 1,732 | -3.183 | 0.288 | -4.248 | -1.701 | 2.548 |
| PC2 | Bump | 1,635 | 0.133 | 1.715 | -10.129 | 10.784 | 20.913 |
| | Pothole | 242 | -2.895 | 2.484 | -10.181 | 3.292 | 13.474 |
| | Normal | 1,732 | 0.279 | 1.847 | -7.742 | 9.968 | 17.710 |
| PC3 | Bump | 1,635 | -0.146 | 1.824 | -9.932 | 10.538 | 20.471 |
| | Pothole | 242 | 3.388 | 2.029 | -2.726 | 8.408 | 11.134 |
| | Normal | 1,732 | -0.335 | 1.508 | -7.834 | 5.971 | 13.805 |

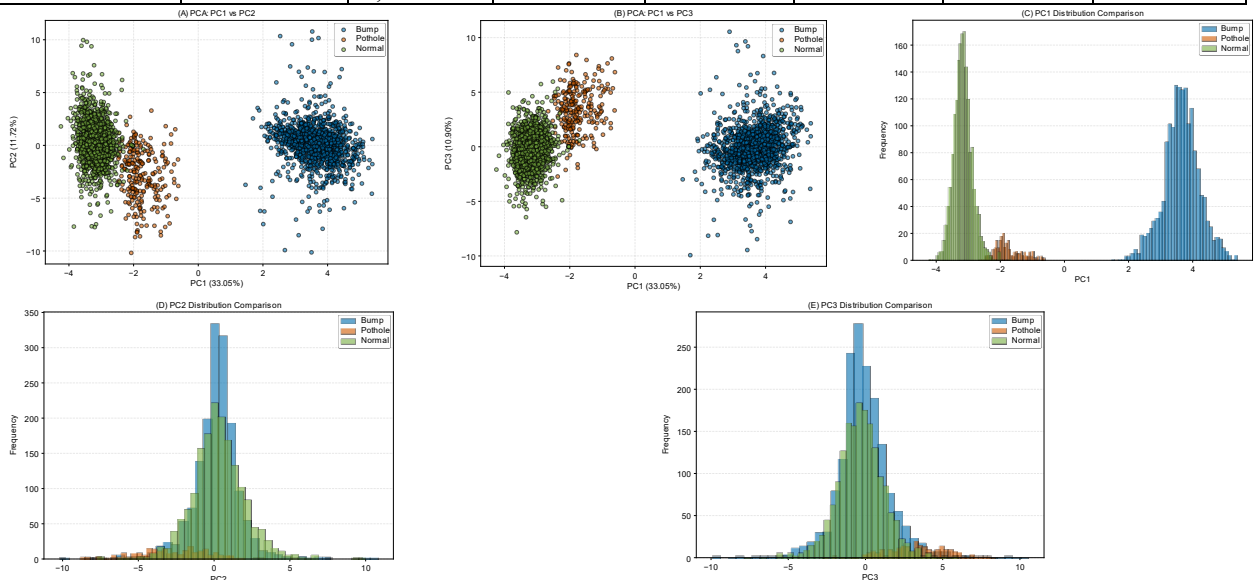

*Figure 15. Principal Component Analysis (PCA) plots that show how different bumps and potholes are from each other. The subplots show the following: (A) PCA1 compared to PCA2, (B) PCA1 compared to PCA3, (C) PCA1 distribution compared to PCA2, (D) PCA2 distribution compared to PCA3, and (E) PCA3 distribution compared to PCA1.*

Cross-correlation analysis indicated strong autocorrelation for bump data (0.681), moderate autocorrelation for potholes (0.619), and low autocorrelation for normal roads (0.430). The boxplots and histograms in **Figure 14** illustrate the distribution and frequency of z-axis accelerations for bumps and potholes, showing clear separation and increasing variability from normal to pothole to bump conditions. Principal component analysis (PCA), visualized in **Figure 15**, further highlights these differences, with PCA scatter plots revealing distinct clustering of bump and pothole events across the first three principal components. Collectively, the table and figures confirm that z-axis accelerometer data exhibits unique signatures for each road condition, emphasizing its effectiveness for accurate road anomaly detection.

# Usage Notes

This comprehensive road anomaly detection dataset provides researchers and practitioners with multimodal data for developing and evaluating intelligent transportation systems. The following guidelines facilitate effective utilization of the dataset.

## Dataset Access and Structure

The Road Anomaly with GIS and Weather Dataset on Figshare (https://doi.org/10.6084/m9.figshare.30341143.v2) has 103 data sessions that are arranged in a hierarchical way to make them easy to access and analyse in multiple ways. As shown in **Figure 9**, the dataset is split into four main parts: Raw Data, Combined CSV with GIS and Weather Data, Isolated Data, and GIS Data. All of the sensor CSV files for that session are in that folder. This includes both calibrated and uncalibrated readings from the accelerometer, gyroscope, magnetometer, barometer, compass, gravity, and GPS sensors, as well as annotation and metadata files. Within each session, there is a separate camera subfolder that contains annotation data and a text file that links to the corresponding video stored on Google Drive. This lets researchers access full recordings without having to manually segment them. The merged CSV files have sensor, GIS, and weather data (temperature, humidity, wind speed, and atmospheric pressure) that are all in sync with each other every 0.01 seconds. The Isolated Data folder also separates normal and anomaly samples so that they can be analysed more closely. The GIS Data folder, on the other hand, has QGIS and elevation files for visualising spatial and topographical data. This well-organised structure makes sure that sensor, video, geographic, and environmental data are all combined in a way that makes them easy to find and use for in-depth research.

## Recommended Applications

The dataset supports: road infrastructure monitoring with automated detection using distinct z-axis variance patterns (bump $\sigma^2 = 1.202$, pothole $\sigma^2 = 1.397$, normal $\sigma^2 = 0.667$ m/s$^{22}$); machine learning with 1,635 bump, 242 pothole, and 1,726 normal samples where PCA captures 55.67% variance in three components; autonomous vehicle development with comprehensive sensor suite mirroring typical AV capabilities; and urban planning with GIS integration enabling spatial analysis across the 170 km Rajshahi route.

## Working with the Data

Both calibrated (bias-corrected, Table 6) and uncalibrated streams are provided. Combined CSV files offer 0.01s temporal alignment for integrated analysis. Video annotations include millisecond timestamps for matching sensor events to 4K (3840×2160, 30fps) frames. Technical validation shows significant condition differences (bump vs. pothole: $t = 3.573$, $p < 0.001$; pothole vs. normal: $t = -3.987$, $p < 0.001$). Allan variance confirms acceptable noise levels, with cross-device validation on Nothing Phone (3a).

## Limitations

Consider: geographic specificity (Rajshahi only, 26.9-31.2°C, 10-24% humidity); sample imbalance (242 potholes vs. 1,635 bumps, 1,726 normal) requiring augmentation strategies; device-specific characteristics (Samsung Galaxy S24, Android 15.0); no video in isolated folders; and 0.01s downsampling in combined files, though original >100 Hz streams remain available.

# Acknowledgments

The authors would like to express their gratitude to ChatGPT, Quill Bot, and Grammarly, among other AI tools, for their invaluable help in polishing and enhancing the writing's lucidity.

# Author Contributions

A.K.: Conceptualization, supervision, project administration, critical revision. D.M.: Conceptualization, supervision, funding acquisition, review & editing. S.G.R.: Methodology, Data collection, preprocessing, experiments, writing—original draft support. F.B.S.: Literature review, visualization, writing—original draft, validation. M.F.A.: Methodology, software, formal analysis, writing—original draft. K.R.: Dataset curation, experiments, validation. M.A.R.: Data curation, implementation support, results verification. M.F.N.: Data collection, preprocessing, statistical checks. M.A.A.: Implementation, visualization, replication experiments. K.K.: Validation, review & editing, technical guidance. P.N.S.: Supervision, Reviewing and editing.

# Funding

# Competing Interests

The authors declare no competing interests

# Data Availability

The dataset supporting this study is publicly available on Figshare at https://doi.org/10.6084/m9.figshare.30341143.v2 [34].

# Code Availability

The code is available at https://github.com/rabbanishaikh/One-Million-Observations-Dataset.git